\documentclass{article}
\usepackage{appendix}
\usepackage{times}
\usepackage{subcaption}

\usepackage{multicol}
\usepackage[bookmarks=true]{hyperref}
\usepackage{amsmath}
\usepackage[capitalise]{cleveref}
\usepackage{xcolor}
\usepackage{graphicx}
\usepackage{booktabs}
\usepackage{wrapfig}
\usepackage{lipsum}
\usepackage{float}
\usepackage{enumerate}
\usepackage[most]{tcolorbox}
\usepackage{lipsum}
\usepackage{adjustbox}
\usepackage{subcaption}
\usepackage{listings}
\usepackage{siunitx}
\usepackage{pifont}
\usepackage[utf8]{inputenc} 
\usepackage{amssymb}
\usepackage{bm}
\usepackage{cuted}
\usepackage{cite}
\usepackage{tikz}
\usepackage{glossaries}
\usepackage{multirow}
\usepackage{makecell}
\newcommand{\tabincell}[2]{\begin{tabular}{@{}#1@{}}#2\end{tabular}}

\newacronym{llm}{LLM}{Large Language Model}
\newacronym{vlm}{VLM}{Vision Language Model}
\newacronym{vla}{VLA}{Vision Language Action}
\newacronym{rag}{RAG}{Retrieval Augmented Generation}
\newacronym{lora}{LoRA}{Low Rank Adaptation}
\newacronym{mpc}{MPC}{Model Predictive Controller}
\newacronym{rmse}{RMSE}{Root Mean Square Error}
\newacronym{hmi}{HMI}{Human Machine Interaction}
\newacronym{ads}{ADS}{Autonomous Driving Systems}
\newacronym{ml}{ML}{Machine Learning}
\newacronym{nn}{NN}{Neural Network}
\newacronym{rl}{RL}{Reinforcement Learning}
\newacronym{ai}{AI}{Artificial Intelligence}
\newacronym{obc}{OBC}{OnBoard Computer}
\newacronym{gpu}{GPU}{Graphics Processing Unit}
\newacronym{cpu}{CPU}{Central Processing Unit}
\newacronym{ram}{RAM}{Random Access Memory}
\newacronym{ros}{ROS}{Robot Operating System}
\newacronym{peft}{PEFT}{Parameter Efficient Fine-Tuning}
\newacronym{cot}{CoT}{Chain of Thought}
\newacronym{sft}{SFT}{Supervised Fine-Tuning}
\newacronym{grpo}{GRPO}{Group Relative Policy Optimization}
\newacronym{eai}{Embodied AI}{Embodied Artificial Intelligence}
\newacronym{sota}{SotA}{State of the Art}
\newacronym{rlvr}{RLVR}{Reinforcement Learning from Verifiable Rewards}
\newcounter{bubblegroup}

\crefname{lstlisting}{Listing}{Listings}
\Crefname{lstlisting}{Listing}{Listings}
\crefname{bubblegroup}{Bubble}{Bubbles}
\Crefname{bubblegroup}{Bubble}{Bubbles}

\definecolor{mpcinstructcolor}{HTML}{EFD0E3}
\definecolor{mpcpromptcolor}{HTML}{DC9EC9}
\definecolor{mpcoutcolor}{HTML}{CA6CAE}
\definecolor{usercolor}{HTML}{CCE4EA}
\definecolor{robocolor}{HTML}{99CAD5}
\definecolor{llmcolor}{HTML}{66AFC0}

\newtcolorbox{narratebubble}{%
  colback=gray!20, 
  colframe=gray, 
  fonttitle=\bfseries,
  sharp corners=south,
  boxrule=0.5mm,
  leftrule=0.5mm,
  rightrule=0.5mm,
  toprule=0.5mm,
  bottomrule=0.5mm,
  title=Human Narration,
  coltitle=white,
  enhanced,
  breakable,
}

\newtcolorbox[use counter=bubblegroup]{mpcbubble}[2]{%
  colback=mpcpromptcolor!20,
  colframe=mpcpromptcolor,
  fonttitle=\bfseries,
  sharp corners=south,
  boxrule=0.5mm,
  leftrule=0.5mm,
  rightrule=0.5mm,
  toprule=0.5mm,
  bottomrule=0.5mm,
  label type = bubblegroup,
  label={bubble:#1}, 
  title=Bubble~#1: MPCxR1 -- #2,
  coltitle=white,
  enhanced,
  breakable,
}

\newtcolorbox[use counter=bubblegroup]{decisionbubble}[2]{%
  colback=robocolor!20,
  colframe=robocolor,
  fonttitle=\bfseries,
  sharp corners=south,
  boxrule=0.5mm,
  leftrule=0.5mm,
  rightrule=0.5mm,
  toprule=0.5mm,
  bottomrule=0.5mm,
  label type = bubblegroup,
  label={bubble:#1}, 
  title=Bubble~#1: DecisionxR1 -- #2,
  coltitle=white,
  enhanced,
  breakable,
}

\newtcolorbox[use counter=bubblegroup]{whichllmbubble}[3]{%
  colback=robocolor!20,
  colframe=robocolor,
  fonttitle=\bfseries,
  sharp corners=south,
  boxrule=0.5mm,
  leftrule=0.5mm,
  rightrule=0.5mm,
  toprule=0.5mm,
  bottomrule=0.5mm,
  label type = bubblegroup,
  label={bubble:#1}, 
  title=Bubble~#1: DecisionxR1 -- #2 -- #3,
  coltitle=white,
  enhanced,
  breakable,
}

\newcommand{\cmark}{\ding{51}}%
\newcommand{\xmark}{\ding{55}} 

\usepackage{amssymb}

\usepackage{listings}
\usepackage{xcolor}

\definecolor{ragblue}{rgb}{0.95,0.98,1}
\lstdefinestyle{ragstyle}{
  backgroundcolor=\color{ragblue},
  basicstyle=\ttfamily\scriptsize,
  breaklines=true,
  frame=single,
  rulecolor=\color{black},
  frameround=tttt,
  captionpos=b,
  showstringspaces=false
}

\usepackage[preprint]{corl_2025} 

\usepackage{eso-pic}

\newcommand\AtPageUpperCenterNotice[1]{%
  \AtPageUpperLeft{%
    \put(\LenToUnit{0.5\paperwidth},\LenToUnit{-2cm}){\makebox[0pt]{#1}}%
  }%
}

\AddToShipoutPictureBG*{%
  \AtPageUpperCenterNotice{%
    \parbox[b][2cm][c]{\paperwidth}{%
      \centering
      \fontsize{12}{14}\selectfont
      \color{gray!50}
      This paper has been accepted for publication at the\\
      Conference on Robot Learning (CoRL), Seoul 2025. 
    }%
  }%
}

\AddToShipoutPictureBG*{%
  \AtPageLowerLeft{%
    \raisebox{20pt}{\makebox[\paperwidth]{\begin{minipage}{21cm}\centering
    \fontsize{10}{10}\selectfont
    \textcolor{gray!50}{
      This work is licensed under a 
      Creative Commons Attribution 4.0 International License (CC BY 4.0).\\
      Please cite the published version when available on the Proceedings of Machine Learning Research (PMLR). 
      For more information, see \url{https://proceedings.mlr.press}.
    }
    \end{minipage}}}%
  }%
}

\title{\vspace{-0.0cm}RobotxR1: Enabling Embodied Robotic Intelligence on Large Language Models through Closed-Loop Reinforcement Learning \vspace{-0.0cm}}

\author{
  Liam Boyle$^{\dagger, *}$, 
  Nicolas Baumann$^{\dagger, \ddagger, *}$, 
  Paviththiren Sivasothilingam$^{\dagger}$,\\
  \textbf{Michele Magno}$^{\dagger}$,
  \textbf{Luca Benini}$^{\ddagger}$\\
  $^{\dagger}$Center for Project-Based Learning, $^{\ddagger}$Integrated Systems Laboratory\\
  ETH Zurich
}

\begin{document}
\maketitle
\footnotetext[1]{Denotes qual contribution. Code at: \href{https://github.com/ForzaETH/LLMxRobot}{github.com/ForzaETH/LLMxRobot}}

\begin{figure}[htb]
    \vspace{-1.0cm}
    \centering
    \begin{adjustbox}{center}
        \includegraphics[width=0.8\textwidth]{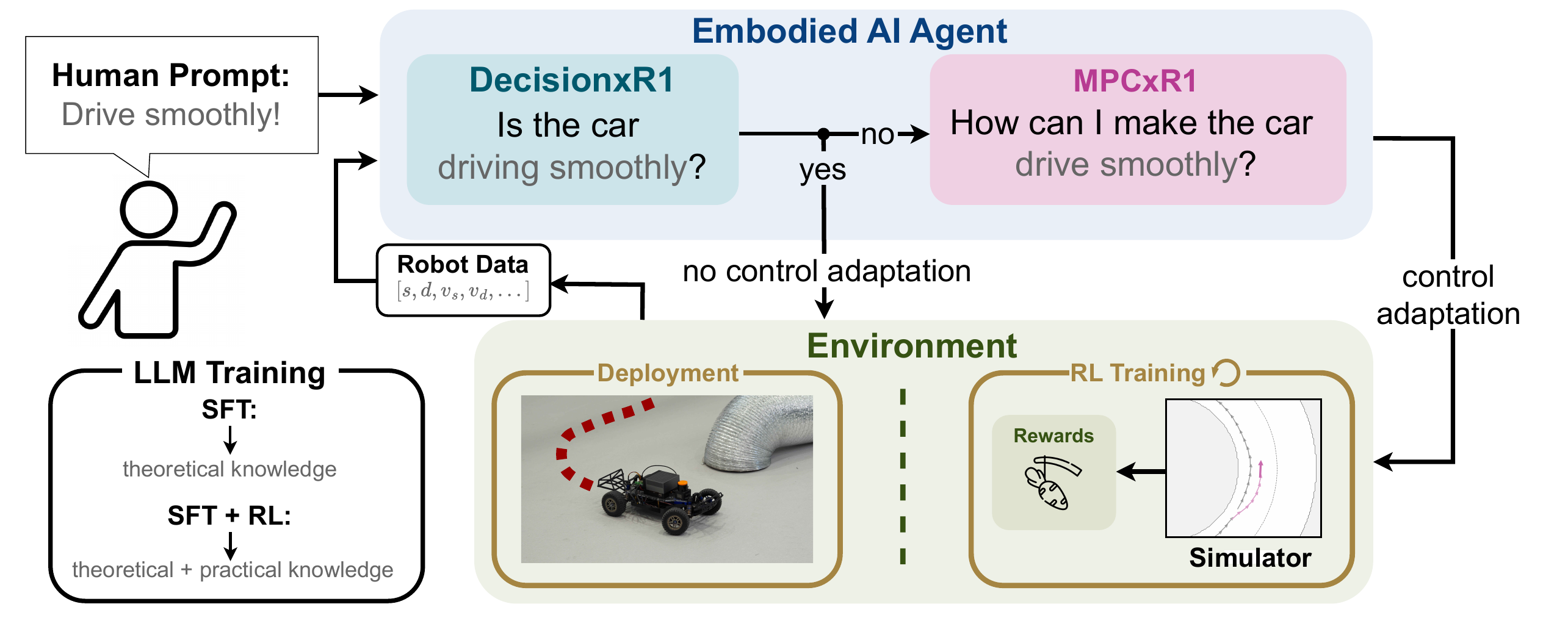}
    \end{adjustbox}
    \caption{Overview of the proposed \acrshort{eai} agent for autonomous driving. The agent consists of a \textit{DecisionxR1} module and an \textit{MPCxR1} module that work in tandem to achieve a user-specific driving behavior. The \acrshortpl{llm} at the core of each module are trained with \acrshort{sft} and \acrshort{rl}. During \acrshort{rl} training, the system interacts with its simulation environment, where it is rewarded for behavioral adherence. The trained agent is deployed on a scaled autonomous car for real-world experiments.}
    \label{fig:title}
\end{figure}
\vspace{-0.0cm}
\glsresetall

\begin{abstract}
    Future robotic systems operating in real-world environments will require on-board embodied intelligence without continuous cloud connection, balancing capabilities with constraints on computational power and memory. This work presents an extension of the \textit{R1-zero} approach, which enables the usage of low parameter-count \glspl{llm} in the robotic domain. The \textit{R1-Zero} approach was originally developed to enable mathematical reasoning in \glspl{llm} using static datasets. We extend it to the robotics domain through integration in a closed-loop \gls{rl} framework. This extension enhances reasoning in \gls{eai} settings without relying solely on distillation of large models through \gls{sft}. We show that small-scale \glspl{llm} can achieve effective reasoning performance by learning through closed-loop interaction with their environment, which enables tasks that previously required significantly larger models. In an autonomous driving setting, a performance gain of 20.2\%-points over the \gls{sft}-based baseline is observed with a \textit{Qwen2.5-1.5B} model. Using the proposed training procedure, \textit{Qwen2.5-3B} achieves a 63.3\% control adaptability score, surpassing the 58.5\% obtained by the much larger, cloud-bound \emph{GPT-4o}. These results highlight that practical, on-board deployment of small \glspl{llm} is not only feasible but can outperform larger models if trained through environmental feedback, underscoring the importance of an interactive learning framework for robotic \gls{eai} --- one grounded in practical experience rather than static supervision.
\end{abstract}

\keywords{Large Language Models, Reinforcement Learning, Embodied AI, Constrained Hardware}


\glsresetall
\section{Introduction}\label{sec:intro}
    The robotics domain has seen significant progress through data-driven \gls{ml} approaches, where the prevailing strategy has been to construct large-scale datasets \cite{caesar2020nuscenes, waymo, o2024openx, coco, laion} to train increasingly large and complex \glspl{nn} \cite{bev_former, openvla, octo_2023, rt-1, paligemma, clip} in a supervised manner. The underlying assumption is that larger datasets provide greater task and environmental diversity, reducing the likelihood that robots encounter unfamiliar scenarios at deployment \cite{clip, openvla}. However, real-world environments are inherently unpredictable and contain numerous edge cases that are intractable to fully capture in any dataset \cite{libero, vlabench}. Humans rely on prior knowledge and contextual reasoning to recognize anomalous situations and adapt their behavior accordingly \cite{wen2023dilu, pavone2024gtc, manip_anything}. To date, \glspl{llm} represent the closest approximation to artificial knowledge-based systems, making their integration into robotic systems a promising direction for \gls{eai} \cite{wen2023dilu, palm_saycan, manip_anything}.

    The recent release of \citet{r1}, introducing the \textit{DeepSeek R1-Zero} method, marks a significant advancement in enabling reasoning within \glspl{llm} \cite{r1,embodiedR}. The approach integrates \gls{rlvr} with \glspl{llm} applied for mathematical datasets such as \texttt{AIME 2024}, \texttt{MATH-500}, or \texttt{GSM8K} \cite{gsm8k}, and has sparked discussion regarding the extent to which such models are capable of solving problems considered to be at PhD level in mathematics \cite{phdlevelmath_doubt1, phdmath_doubt0}. However, this required training a massive \gls{llm} within a \gls{rl} loop --- an approach that incurs extremely high computational costs \cite{r1, unsloth, tinyzero}. In this work, we focus on \gls{ads} as a concrete and high-stakes example of embodied robotic intelligence, where deploying such large models poses a major limitation, as it prevents edge deployment without reliance on cloud infrastructure. Specifically for \gls{ads} and other robotic domains, ubiquitous cloud connectivity is infeasible and introduces serious security vulnerabilities \cite{iot_good0, iot_good1, slm}. 

    One potential solution is to distill the reasoning capabilities of large \glspl{llm} into smaller, more efficient models through \gls{sft}, an approach also demonstrated effectively in the \textit{R1-Zero} framework \cite{r1}. 
    Distilled models may inherit some reasoning ability, but unlike human reasoning, which is tightly coupled with sensory feedback, situational awareness, and physical context, these models operate in abstraction, lacking the closed-loop perception-action embodiment that underpin robust robotic intelligence \cite{silverwelcome}. Moreover, \gls{rlvr} was first introduced to operate on static math or coding datasets \cite{shao2024deepseekmath, r1}, whereas \gls{rl} in robotics typically involves interaction with dynamic environments \cite{massive_parallel_rl, sophy, silverwelcome}. Success in solving abstract, mathematical problems does not necessarily translate to the embodied, context-dependent reasoning needed for robotic operation. 
    Hence, we argue that the type of reasoning required for robotics is fundamentally different from that assessed in PhD-level mathematics benchmarks. Similarly, one must ask whether \gls{ads} should rely on models trained to mimic mathematical problem-solving, or rather on systems that acquire reasoning capabilities through direct interaction with their environment.
    

    Our contributions are fourfold:  
    \textbf{(i)} \textit{RobotxR1} is introduced as an extension of the \textit{R1-Zero} framework, showing that  direct interaction of \glspl{llm} with their environment via closed-loop \gls{rl} is possible in an \gls{ads} setting, moving beyond static dataset \gls{rlvr} training. 
    \textbf{(ii)} Edge-deployable \glspl{llm} are shown to effectively learn through interaction with a scaled autonomous vehicle, with a 3B model achieving a 63.3\% control adaptability score --- surpassing the 58.5\% obtained by the much larger cloud-based \textit{GPT-4o}.  
    \textbf{(iii)} Interaction-based training is shown to substantially benefit small models, with a 1.5B model achieving a 20.2\%-point improvement over its \gls{sft}-only baseline, underscoring the value of embodied learning through interaction, in robotic \gls{eai}.
    \textbf{(iv)} The proposed method has low computational demands, with training feasible on a single consumer-grade \gls{gpu} (e.g., RTX 3090), and with the \textit{Qwen2.5} 1.5B and 3B models being capable of deployment on an embedded Jetson Orin AGX serving as the robot's \gls{obc}.


\section{Related Work} \label{sec:rw}

\paragraph{Robot Control and LLMs:}
Recent research has explored the integration of \glspl{llm} with robotic control systems \cite{rsspaper, l2r, ismail2024narrate, ma2023eureka, wen2023dilu, palm_saycan}. These works consistently find that \glspl{llm} are ill-suited for direct low-level control, and instead should influence behavior indirectly, typically by shaping a reward or cost signal for a dedicated low-level controller, such as in a \gls{mpc} framework \cite{rsspaper, l2r}. \citet{ismail2024narrate} propose using an \gls{llm} to generate objective functions and constraints for manipulation tasks conditioned on human prompts. Their architecture combines the flexibility of \glspl{llm} with established and classic \gls{mpc} controllers.
Similarly, \citet{rsspaper} introduces an open-source framework for \gls{llm}-based robot interaction through a low-level \gls{mpc}, and demonstrates it on a 1:10 scaled autonomous racing car. Our work targets the same tightly constrained deployment scenario. However, existing work primarily relies on \gls{sft} or closed-source, cloud-based models like \textit{GPT-4}, which cannot be finetuned locally and are thus limited to prompt engineering. As such, these methods lack trainable \gls{eai} reasoning capability.

\paragraph{Reasoning with LLMs:}
The \textit{R1-zero} framework introduced by \citet{r1} demonstrates how an \gls{llm} can be integrated into a \gls{rlvr} loop for structured reasoning tasks on static datasets, such as multiple-choice math questions \cite{shao2024deepseekmath}. The reward signal combines formatting and correctness objectives, where formatting ensures the answer can be reliably parsed. This approach has shown \gls{sota} reasoning performance in mathematics-based tasks. Follow-up work includes \citet{dao2025alphamaze}, which applies \gls{rlvr}, and more specifically \gls{grpo}, in an \textit{R1-zero} fashion to enable spatial reasoning in maze-solving tasks, and \citet{azzolini2025cosmos}, which extends the framework to a \gls{vlm} for reasoning in embodied agents within autonomous driving scenarios. However, these methods are limited to static datasets and lack interaction with dynamic environments, missing out on streams of experiences that arise from closed-loop environmental feedback \cite{silverwelcome}.

By contrast, \gls{rl} in robotics typically involves continuous interaction with an environment, forming a closed feedback loop. In this work, we close this gap by adapting the \textit{R1-zero} process to a dynamic \gls{ads} simulation environment, where the \gls{llm} interfaces with a low-level \gls{mpc} controller and must reason over human driving commands to guide the robot's behavior. To the best of our knowledge, this is the first \gls{llm}-driven training loop in continuous control with direct simulation feedback, enabling real-time, language-grounded decision-making combined with classical control.

\section{Methodology} \label{sec:method}
 This work leverages the publicly available framework introduced in \cite{rsspaper} to enable a comparison with purely \gls{sft}-trained \glspl{llm}, while significantly extending the \emph{DecisionxLLM} and \emph{MPCxLLM} architectures of \cite{rsspaper} to support robotic reasoning in dynamic environments. The existing \gls{rag} structure, including the use of five retrieved memories as in the baseline \cite{rsspaper}, and the proposed \gls{lora}-based \gls{sft} method are leveraged. In this work, the \emph{Qwen} family of models \cite{bai2023qwen}, ranging from 1.5 to 7B parameters, is adopted as the primary \gls{llm} architecture, motivated by the performance reported in \cite{rsspaper}. The \emph{R1-zero} framework is integrated into this setup, resulting in \textit{RobotxR1}, which encompasses the \emph{DecisionxR1} and \emph{MPCxR1} modules, to enable full \gls{rl}-based embodiment of the underlying \glspl{llm} within a holistic pipeline that follows sequential decision making and control adaptation, as illustrated in \Cref{fig:title}.

\subsection{Robotic Autonomy Stack and MPC}\label{subsec:stack}
 The utilized robotic autonomy stack follows the 1:10 scaled racecar \cite{forzaeth}, where the racing line corresponds to a minimum curvature trajectory computed for a closed race track. This racing line is then tracked utilizing a kinematic \gls{mpc} defined in a curvilinear coordinate system as in \cite{rsspaper}. The robot state is defined by $x=\begin{bmatrix} s &n &\Delta \phi & \delta & v \end{bmatrix}^{T}$, where \( s \) and \( n \) represent the longitudinal and lateral distances to the racing line; \( \Delta \phi \) denotes the heading error relative to the racing line; \( \delta \) is the steering angle; and \( v \) is the longitudinal velocity. The MPC control input is $u = [\Delta \delta, a]^{T}$.

 \begin{equation}\label{eq:mpc}
    \min_u \ J(x, u) = \sum_{i=0}^{N-1} q_n n_i^2 + q_v (v_i - v_\text{ref})^2 +q_{\alpha} \Delta \phi_i^2 +\Vert \Delta u_i \Vert_{q_R} \quad 
    s.t. \quad
    x_{i}\in \mathcal{X}, u_{i}\in \mathcal{U}
\end{equation}

Here, $N$ denotes the prediction horizon. The weights $q_n$, $q_{\alpha}$, and $q_v$ correspond to state terms, while $q_R$ is a diagonal matrix penalizing control inputs. State and input constraints are defined by $\mathcal{X}$ and $\mathcal{U}$. Velocity $v$ and steering angle $\delta$ are bounded in magnitude, and lateral error $n$ is constrained within track boundaries, modulated by an inflation factor $\epsilon$ for safety. Steering rate and acceleration are also constrained. The default \gls{mpc} parameters have been empirically obtained and optimized to track the racing line effectively.

The cost function weights are exposed as online-tunable parameters, enabling real-time adaptation of driving behavior. Additionally, constraints such as for example the boundary inflation factor $\epsilon$ and the velocity bounds can be modified by the \gls{llm}, allowing the robot’s driving policy to be dynamically adjusted in response to natural language instructions (more information in \Cref{app:rag}).

\subsection{DecisionxR1 --- Enhanced Reasoning}\label{subsec:decisionxr1}
\begin{wrapfigure}{R}{0.58\columnwidth}
    \vspace{-10pt}
    \centering
    \includegraphics[trim={0 0.3cm 0 0.3cm},clip,width=0.58\columnwidth]{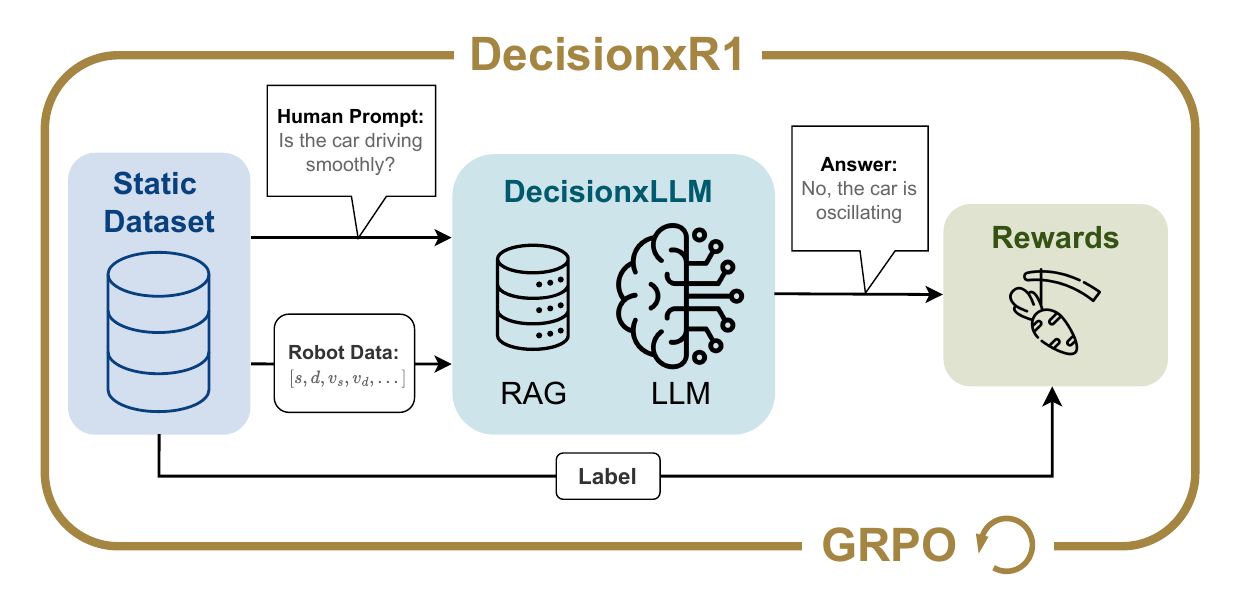}
    \caption{The DecisionxR1 module consists of a \gls{rag} enhanced \gls{llm} that uses robot state information to determine if the car is adhering to the desired behavior prompted by the human. Each answer of the \gls{llm} is assigned a format and correctness reward that are used to train the model with \gls{grpo}.}
    \label{fig:decision_llm}
    \vspace{-10pt}
\end{wrapfigure}

The objective of the \textit{DecisionxR1} module, depicted in \Cref{fig:decision_llm}, is to teach an \gls{llm} to reason about the current driving behavior of an autonomous racing car. More specifically, the \gls{llm} should decide whether the car is adhering to the driving behavior prompted by the human using a given history of robot states. Following the recent successes of other works \cite{dao2025alphamaze, azzolini2025cosmos, rl4vlm, apple_llarp} on training \glspl{llm} for embodied reasoning with \gls{rl}, we adopt the two-stage \emph{R1-zero} training procedure introduced in \cite{r1} to finetune the \textit{DecisionxR1} module with \gls{rlvr}. In the first stage, the model undergoes \gls{sft} following the methodology of \cite{rsspaper}, which distills embodiment-specific knowledge into pre-trained \textit{Qwen} 1.5B and 3B models. In the second stage, the model is further optimized using \gls{rlvr}, more specifically \gls{grpo}, guided by a static decision-making dataset and reward functions designed to reinforce both decision accuracy and output structure.

\textbf{Decision-Making Dataset:} Similar to \cite{r1, dao2025alphamaze}, we build a static dataset in which each instance is formulated as a binary classification task. The dataset was obtained by driving the robot in simulation in eight different driving styles (e.g., centerline tracking, reversing, raceline tracking, etc.), and recording robot state information $x$, as in \Cref{eq:mpc}. The binary behavior adherence label is computed programmatically. Programmatic labeling is done by defining a set of rules in $\mathcal{B}$, that evaluate how closely the robot’s actions match a target driving style. For example, for reversing, if the longitudinal velocity \( v \) is negative, the label is set to 1 (adherence), otherwise 0 (non-adherence).

\textbf{Reward Modeling for Decision Making:} We employ two kinds of rewards, which we roughly define as accuracy and formatting rewards, such that the total reward can be described as
\begin{equation}\label{eq:decision_reward}
    R_{\text{DecisionxR1}} = R_{\text{accuracy}} + R_{\text{fmt}},
\end{equation}
where $R_{\text{accuracy}}$ is the reward the model gets for correctly determining if the robot is adhering to the prompted behavior and $R_{\text{fmt}}$ is a formatting reward that should encourage the model to structure its answer into reasoning, explanation, and answer sections (e.g. using \texttt{<reasoning>} and \texttt{</reasoning>} tags to mark the beginning and end of the reasoning section). While the correctness reward incentivizes accurate decision-making, the formatting reward serves a dual purpose: it encourages the model to articulate its reasoning and ensures that the output remains structured and easily parsable.

\subsection{MPCxR1 --- Enhanced Control Adaptability}
\begin{figure}[htb]
    \vspace{-0.25cm}
    \centering
    \includegraphics[trim={0 0 0 0},clip,width=0.9\linewidth]{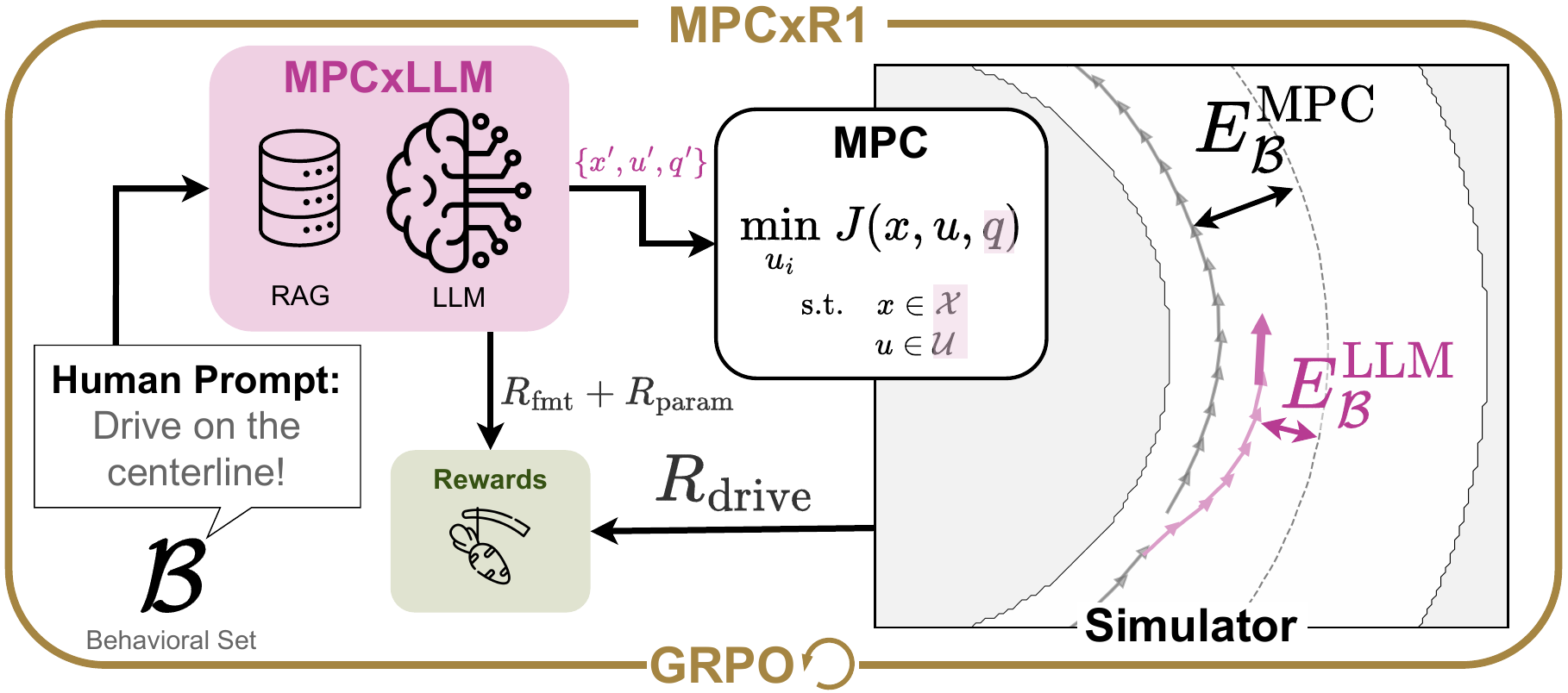}
    \caption{Schematic overview of the proposed \textit{MPCxR1} training procedure. The \gls{llm} leverages \gls{rag} context, following \cite{rsspaper}, to generate \gls{mpc} parameters tailored to the desired driving behavior $\mathcal{B}$. The \gls{llm} and \gls{mpc} operate in a closed-loop with a simulator, which computes both the baseline \gls{rmse} under default \gls{mpc} parameters ($E^{\text{MPC}}_{\mathcal{B}}$) and the \gls{rmse} resulting from the \gls{llm}-generated parameters ($E^{\text{LLM}}_{\mathcal{B}}$), both relative to the behavioral objective, later introduced in \Cref{eq:mpc}. These are used to compute the behavioral reward $R_{\text{drive}}$, while the \gls{llm}'s textual output is evaluated for formatting ($R_{\text{fmt}}$) and parameter validity ($R_{\text{param}}$). This framework extends \emph{R1-zero} from static dataset training to a fully embodied \gls{rl} setup via \gls{grpo}.}
    \label{fig:mpc_llm}
\end{figure}

In contrast to the \textit{DecisionxR1} module introduced in \Cref{subsec:decisionxr1}, control adaptability in robotic systems cannot rely solely on static datasets. As in many robotic applications, effective behavior emerges through interaction with the environment rather than from parameters learned via \gls{sft} alone. As an analogy, this distinction reflects the difference between learning to drive by reading a manual (\gls{sft}) and learning through actual driving lessons (\gls{rlvr}). In this setting, the \textit{MPCxR1} module places the \gls{llm} in a closed-loop with the \gls{mpc}, enabling it to influence control behavior based on interaction, thereby facilitating embodied learning and control adaptation through experience. 

\begin{wrapfigure}{R}{0.5\textwidth}
\vspace{-10pt}
\begin{minipage}{0.5\textwidth}
\small
\begin{mpcbubble}{1}{Example Prompt}
Adapt the tuneable parameters of the MPC so that the car achieves the following: \textbf{\textit{"Drive at 1.83 m/s as closely as possible"}}.\\
This is the MPC formulation: \texttt{MPC Formulation}.\\
Return format:
\vspace{-0.15cm}
\begin{verbatim}
new_mpc_params = {
    param1: new_value1,
    ...
}
\end{verbatim}
\end{mpcbubble}
\setcounter{bubblegroup}{1}
\end{minipage}
\vspace{-10pt}
\end{wrapfigure}

\textbf{Closed-Loop RL Environment:} As illustrated in \Cref{fig:mpc_llm}, the standard \textit{R1-zero} training process is modified to incorporate closed-loop feedback from a driving simulation. The training spans a range of driving behaviors $\mathcal{B}$ (see \Cref{app:thebss} for the full set), with each prompt specifying a distinct behavioral objective. One such case is instructing the \textit{MPCxR1} module to \emph{"Drive at 1.83 m/s as closely as possible"}, as shown in \Cref{bubble:1}. In response, the \gls{llm} generates \gls{mpc} parameters intended to realize the desired behavior. During training, these \gls{mpc} parameters are tested in a closed-loop simulation such that behavior adaptation rewards can be computed. Further, to enhance and emphasize the generalization capabilities of the \gls{rlvr}-trained \gls{llm}, the \gls{rl} training is conducted on a simplistic circle map, while being evaluated on a complex racing track (see the track layouts in \Cref{app:maps}).

\textbf{Reward Modeling for Control Adaptability:} To train the \textit{MPCxR1} module, we use three different rewards, a driving, a formatting, and a parameter extraction reward, that feed into the total reward, $R_{\text{MPCxR1}}$ as follows:
\begin{equation}\label{eq:mpc_rew}
    R_{\text{MPCxR1}} = R_{\text{drive}} + R_{\text{fmt}} + R_{\text{param}}, \quad
    R_{\text{drive}} = \max\left( \frac{E_{\mathcal{B}}^{\text{MPC}} - E_{\mathcal{B}}^{\text{LLM}}}{E_{\mathcal{B}}^{\text{MPC}}},\ -4 \right)
\end{equation}
The aim of $R_{\text{drive}}$, is to reward the \gls{llm} for producing \gls{mpc} parameters that result in the correct driving behavior requested by the human prompt. At each training step, these parameters are applied to the \gls{mpc} and the vehicle completes a lap in simulation. Subsequently, the performance can be evaluated using behavior-specific \gls{rmse} metrics, computed for both the default \gls{mpc} and the \gls{llm} adjusted \gls{mpc}. We denote these errors by $E^{\text{MPC}}_{\mathcal{B}} \in [0, \infty)$ and $E^{LLM}_{\mathcal{B}} \in [0, \infty)$, respectively. The driving reward $R_{\text{drive}} \in [1, -4]$ is then computed as the relative improvment achieved by the \gls{llm} over the default \gls{mpc} parameters as shown in \Cref{eq:mpc_rew}. Each behavioral prompt has a corresponding \gls{rmse} formulation, enabling consistent reward computation across tasks.
$R_{\text{fmt}}$ is a formatting reward similar to that in \Cref{subsec:decisionxr1}, encouraging the model to structure its response and reason about its answer. Finally, $R_{\text{param}}$ is a parameter extraction reward that should discourage the \gls{llm} from producing hallucinated or invalid \gls{mpc} parameters, which would result in extraction failures. This composite reward guides the \gls{llm} through \gls{grpo} to generate interpretable, valid, and behavior-aligned control adaptations.

\section{Experiments and Results} \label{sec:result}

\subsection{Training Details}
\label{sec:train_details}
All \gls{rlvr} training was done using \gls{grpo} adapted from the \gls{lora}-based \texttt{unsloth} implementation \cite{unsloth} and executed on consumer-grade \glspl{gpu} (RTX 4070 Ti, RTX 3090). While \gls{lora}-based \gls{grpo} significantly reduces VRAM requirements compared to the original \emph{DeepSeek R1} implementation, it remains significantly more computationally intensive than standard \gls{lora} \gls{sft} training. As a result, no models larger than \emph{Qwen2.5-3B} could be trained with \gls{grpo}, requiring approximately 11 GB of VRAM. The \emph{Qwen2.5-7B} model, used in the baseline of \cite{rsspaper}, was limited to \gls{sft}-only training, while both 1.5B and 3B models were also \gls{sft}-trained for comparison, following the procedure in \cite{rsspaper}. \gls{grpo} training was performed for 750 steps, resulting in an approximately 24 h-long training for the 3B model. For more details on the training procedure, refer to \Cref{app:training_curves}.

\subsection{Decision Making Results}
\begin{wraptable}{R}{0.4\columnwidth}
    \vspace{-10pt}
    \centering
    \begin{adjustbox}{max width=0.4\columnwidth}
    \begin{tabular}{l|c|c|c}
    \toprule
    \textbf{LLM} & \textbf{SFT} & \textbf{RLVR} & \textbf{Accuracy [\%]}\bm{$\uparrow$}\\
    \midrule
    GPT4o        & \xmark & \xmark & \textbf{92.48} \\
    Qwen2.5-1.5B & \xmark & \xmark & 47.09\\
    Qwen2.5-3B   & \xmark & \xmark & 61.24\\
    Qwen2.5-7B   & \xmark & \xmark & 82.47\\
    \midrule
    Qwen2.5-1.5B & \cmark & \xmark & 68.80\\
    Qwen2.5-3B   & \cmark & \xmark & 83.78\\
    Qwen2.5-7B   & \cmark & \xmark & \textbf{87.32} \\
    \midrule
    Qwen2.5-1.5B & \xmark & \cmark & 64.17\\
    Qwen2.5-3B   & \xmark & \cmark & \textbf{78.05}\\
    \midrule
    Qwen2.5-1.5B & \cmark & \cmark & 82.83\\
    Qwen2.5-3B   & \cmark & \cmark & \textbf{86.82}\\
    \bottomrule
    \end{tabular}
    \end{adjustbox}
    \caption{Decision-making accuracy across \glspl{llm} sizes, illustrating the impact of \gls{sft}, and \gls{rlvr}.}
    \label{tab:decision_res}
    \vspace{-10pt}
\end{wraptable}


\Cref{tab:decision_res} compares the decision-making accuracy of baseline models with \gls{sft} and \gls{rlvr} finetuned models. In this comparison, all models, including \textit{GPT-4o}, are augmented with the same \gls{rag} data (more information in \Cref{app:rag}). To ensure comparability, we adopt the same evaluation procedure as in \cite{rsspaper}. The test set comprises 8 human prompts, each querying adherence to a specific driving behavior (e.g., “Is the car driving on the racing line?”). For each behavior, the dataset includes 25 robot state histories exhibiting that behavior, resulting in a total of 200 state trajectories. By pairing each of the 8 prompts with all 200 trajectories, a comprehensive test set of 1600 prompt–trajectory combinations is constructed. 

When finetuning models with \gls{sft}, we observe that the 1.5B, 3B, and 7B models achieve higher decision accuracies than their original counterparts, where the accuracies increase by 21.71\%, 22.54\%, and 4.85\%-points for 1.5B, 3B, and 7B, respectively.

Due to the high memory demands of \gls{grpo}, training 7B models was not feasible on our training infrastructure (see \Cref{sec:train_details}). When 1.5B and 3B models are finetuned solely with \gls{rlvr}, we observe that it is less effective in improving decision-making accuracy compared to their \gls{sft}-only counterparts, achieving an improvement compared to the original base models of 17.08\% and 16.81\%-points for the 1.5B and 3B models. 



 Finally, we evaluate a two-stage training process using \gls{sft} as a pretraining step before applying \gls{rlvr}. This boosts decision accuracy when compared against \gls{sft}-only trained counterparts. For \textit{Qwen2.5-1.5B} by 14.03\%-points (\gls{sft}: 68.8\% / \gls{sft}+\gls{rlvr}: 82.83\%) and for \textit{Qwen2.5-3B} by 3.04\%-points (\gls{sft}: 83.78\% / \gls{sft}+\gls{rlvr}: 86.82\%) over \gls{sft}-only training.

\subsection{Control Adaptability Results}\label{subsec:control_res}
To quantitatively evaluate the impact of a fully embodied training procedure with closed-loop simulator interaction on the \gls{llm}, we assess control adaptability by measuring how well the robot's behavior aligns with human prompts. The agent is compared against the default \gls{mpc} behavior (which is to track the racing line), allowing us to compute a control adaptability improvement score using the open-source \textit{F1TENTH} simulator \cite{babu2020f1tenth, forzaeth}. \gls{sft} training, by distilling \emph{GPT-4o} outputs on the smaller \glspl{llm}, follows the procedure outlined in \cite{rsspaper}, and all evaluations are conducted using the same \gls{rag} context to ensure comparability, see \Cref{app:rag} for further details.

In contrast to the \textit{MPCxR1} training phase, where prompts are sampled from the behavioral set $\mathcal{B}$, each behavior in $\mathcal{B}$ was rephrased and randomized five times to construct the evaluation set $\mathcal{B}'$. Additionally, the evaluation environment differs from the training setup: while \textit{MPCxR1} was trained on a simple circular track, evaluation is conducted on a more complex and realistic track layout (see \Cref{app:maps}). This setup enables testing of generalization by evaluating the model's ability to interpret a randomized behavioral set $\mathcal{B}'$ (more information in \Cref{app:thebss}), highlighting the open-vocabulary capabilities of the \gls{llm}, and to transfer behavior learned on a simple circular training track to a more complex, realistic environment.

\begin{table*}[!htb] 
    \centering 
    \resizebox{\textwidth}{!}{
    \begin{tabular}{l|c|c|c|c|c|c||c|c}
    \toprule
    \textbf{LLM} & \textbf{SFT} & \textbf{RLVR} & \bm{$E_C [m]\downarrow$} & \bm{$E_V [ms^{-1}]\downarrow$} &  \bm{$E_R [ms^{-1}]\downarrow$} & \bm{$E_S [ms^{-2}]\downarrow$} & \textbf{Ext. Fail [\#]}\bm{$\downarrow$} & \textbf{Improve [\%]}\bm{$\uparrow$} \\
    \midrule
    MPC (default) & - & - & 0.68 & 1.98 & 5.44 & 1.77 & - & - \\
    \midrule
    GPT-4o & \xmark & \xmark & 0.65 (5.0\%) & \textbf{0.14 (93.2\%)} & \textbf{0.12 (97.8\%)} & \textbf{1.10 (38.0\%)} & \textbf{0} & \textbf{58.5\%} \\
    Qwen2.5-1.5B & \xmark & \xmark & 0.88 (-29.9\%)$\dagger$ & 0.90 (54.5\%)$\dagger$ & 2.55 (53.1\%) & 1.76 (0.6\%) & 3 & 19.6\%$\dagger$ \\
    Qwen2.5-3B & \xmark & \xmark & \textbf{0.62 (8.2\%)} & 1.51 (23.6\%) & 0.14 (97.5\%) & 1.61 (9.5\%) & \textbf{0} & 34.7\% \\
    Qwen2.5-7B & \xmark & \xmark & 0.68 (0.3\%) & 0.23 (88.2\%) & 0.23 (95.7\%) & 1.43 (19.3\%) & \textbf{0} & 50.7\% \\
    \midrule
    Qwen2.5-1.5B & \cmark & \xmark & \textbf{0.49 (28.1\%)} & 1.01 (49.0\%) & 0.32 (94.2\%) & 4.01 (-126.1\%) & \textbf{0} & 11.3\% \\
    Qwen2.5-3B & \cmark & \xmark & 0.67 (1.9\%)$\dagger$ & 0.43 (78.4\%) & 0.13 (97.6\%) & 1.36 (23.6\%) & 1 & 50.4\%$\dagger$ \\
    Qwen2.5-7B & \cmark & \xmark & 0.53 (21.5\%) & \textbf{0.24 (87.9\%)} & \textbf{0.11 (97.9\%)} & \textbf{1.34 (24.5\%)} & \textbf{0} & \textbf{58.0\%} \\
    \midrule
    Qwen2.5-1.5B & \xmark & \cmark & 0.62 (8.5\%) & 1.89 (4.3\%)$\dagger$ & 3.70 (32.0\%)$\dagger$ & 1.89 (-6.7\%) & 3 & 9.6\%$\dagger$ \\
    Qwen2.5-3B & \xmark & \cmark &  \textbf{0.45 (33.3\%)} & \textbf{0.39 (80.2\%)} & \textbf{0.2 (96.3\%)} & \textbf{1.44 (18.8\%)} & \textbf{0} & \textbf{57.2\%} \\
    \midrule
    Qwen2.5-1.5B & \cmark & \cmark & 0.62 (8.5\%) & 1.05 (47.0\%) & 0.72 (86.7\%) & 1.36 (23.6\%) & \textbf{0} & 41.5\% \\
    Qwen2.5-3B & \cmark & \cmark &  \textbf{0.41 (39.9\%)} & \textbf{0.19 (90.2\%)} & \textbf{0.48 (91.2\%)} & \textbf{1.21 (31.8\%)} & \textbf{0} & \textbf{63.3\%} \\
    \bottomrule
    \end{tabular}
    }
    \caption{Quantitative comparison of \gls{llm} control adaptation through \gls{mpc} interactions, evaluated on \textit{The Grand Tour} map (\Cref{app:maps}). Metrics include deviation from the centerline ($E_C$ [m]), reference velocity tracking error ($E_V$ [ms$^{-1}$]), reversing error ($E_R$ [ms$^{-1}$]), and driving smoothness ($E_S$ [ms$^{-2}$]). \gls{mpc} parameter extraction failures (Ext. Fail [\#]) are where the \gls{llm} output could not be parsed due to invalid parameters. Percentage improvements are computed relative to the baseline \gls{mpc}, which tracks the racing line. The improvement column aggregates overall performance across all metrics. Each entry represents the mean over five independent runs using different prompts in $\mathcal{B'}$. $\dagger$ indicates averages computed excluding extraction failures.}
    \label{tab:control_res}
\end{table*}

As visible from \Cref{tab:control_res}, if neither \gls{sft} nor \gls{rlvr} is applied (i.e., the original models are used), control adaptability improves with model size. \textit{GPT-4o} performs the best at 58.5\%, closely followed by \textit{Qwen2.5-7B} at 50.7\%, both showing strong adaptability. However, \textit{Qwen2.5-1.5B} fails in 3 instances due to unparseable or hallucinated \gls{mpc} parameters, which is detrimental for deployment on a robot.

Among local \gls{sft}-only trained models, solely \textit{Qwen2.5-7B} achieves a comparable improvement (58\%), on par with \textit{GPT-4o}. In contrast, \textit{Qwen2.5-1.5B} achieves only 11.3\%, and while \textit{Qwen2.5-3B} reaches 50.4\%, it suffers from parameter extraction failures, which would result in complete disobedience of robotic behavior adaptation. Evaluating \gls{rlvr} alone, the 7B model cannot be trained due to computational constraints. The 1.5B model performs worse than under \gls{sft}, with improvement dropping from 11.3\% to $9.5\%\dagger$ and three extraction failures. Conversely, the 3B model benefits from \gls{rlvr}, improving from 50.4\% to 57.2\%, a gain of 6.8\%.

When combining \gls{sft} and \gls{rlvr}, both \textit{Qwen2.5-1.5B} and \textit{Qwen2.5-3B} show substantial gains --- 41.5\% and 63.3\%, respectively --- with no extraction failures. Compared to \gls{sft}-only training (analogous to learning from a driving manual), the combined \gls{sft} pretraining and closed-loop \gls{rlvr} training (analogous to studying the manual and taking real driving lessons) yield a 20.2\%-point improvement for \textit{Qwen2.5-1.5B} (\gls{sft}: 11.3\% / \gls{sft}+\gls{rlvr}: 41.5\%) and a 12.9\%-point improvement for \textit{Qwen2.5-3B} (\gls{sft}: 50.4\%\textsuperscript{$\dagger$} / \gls{sft}+\gls{rlvr}: 63.3\%), with \gls{sft}-only still showing critical extraction failures $\dagger$, so the improvment would arguably be even higher.



These findings indicate that a 3B model can outperform even the cloud-based \textit{GPT-4o} --- despite its significantly larger parameter count --- in an embodied robotic setting. While \textit{GPT-4o} achieves a 58.5\% improvement in control adaptability, our proposed training procedure reaches 63.3\% with the \textit{Qwen2.5-3B} model. These results demonstrate that \glspl{llm} can learn to interact with robots through \gls{rl}, with performance gains attributed to the embodied nature of the proposed closed-loop training.

\subsection{Deployment on Physical Robot}
\begin{wrapfigure}{R}{0.5\columnwidth}
    \vspace{-10pt}
    \centering
    \includegraphics[width=0.5\columnwidth]{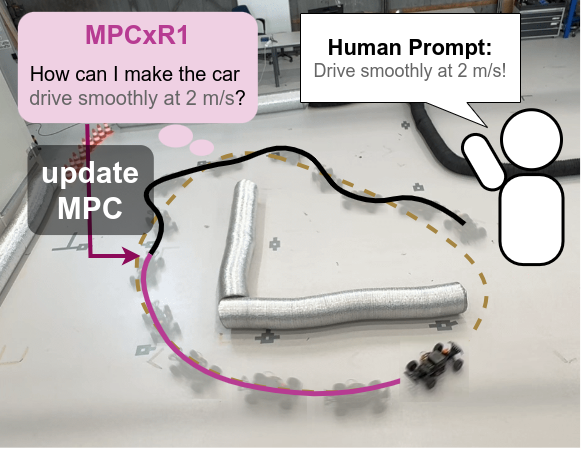}
    \caption{Adaptation of robot behavior in response to user prompts during embedded deployment with the proposed \gls{rlvr} and \gls{sft} trained and \texttt{Q5\_k\_m} quantized \emph{Qwen2.5-3B}.}
    \label{fig:car_experiment}
    \vspace{-10pt}
\end{wrapfigure}
In this experiment, the proposed system is deployed on a physical 1:10 scaled open-source autonomous racing car \cite{forzaeth} running a \texttt{Q5\_k\_m} quantized  \textit{Qwen2.5-3B} model on a Jetson Oring AGX \gls{obc} inferred through \texttt{llama.cpp} \cite{llamacpp}, with a full computational analysis in \Cref{app:compute}. The deployed embedded model achieves a mean throughput of $\sim$38.78 tokens/s and an average latency of $\sim$\SI{8.3}{\second}. The robot is initially placed in a perturbed state, where a human operator has manually altered the \gls{mpc} parameters to induce unstable oscillations. The \textit{MPCxR1} module is then prompted with the instruction: \emph{"Drive smoothly at 2 m/s"}. As observed in \Cref{fig:car_experiment}, the robot's behavior adapts to
satisfy both objectives simultaneously, demonstrating the ability to handle compound instructions, leveraging the open-vocabulary skillset of \glspl{llm}. A second prompt, \emph{"Reverse the car"}, followed by a final prompt, \emph{"Drive normal again"}, are succesfully executed (not visible in \Cref{fig:car_experiment}, but verifiable in \Cref{app:car_bubbles}).


\section{Conclusion}\label{sec:conclusion}
This work presents an \gls{rl}-based training procedure that enables \glspl{llm} in robotic \gls{eai} settings to acquire behaviors through \emph{"learning by doing"}, rather than relying on static datasets or solely \gls{sft}. The proposed method demonstrates that closed-loop interaction can be crucial for small models to successfully learn embodied behaviors. The approach is evaluated in an \gls{ads} setting and enables compact \glspl{llm} to adapt robotic control behavior effectively --- capabilities previously limited to larger models. Using this framework, a \textit{Qwen2.5-1.5B} model achieves up to a 20.2\%-point improvement over \gls{sft}-only training, while the \textit{Qwen2.5-3B} model reaches a 63.3\% control adaptability score, surpassing the performance of the much larger cloud-based \emph{GPT-4o} (58.5\%). These results highlight the viability of feedback-driven learning for \glspl{llm}, enabling compact models in robotic \gls{eai} settings through closed-loop \gls{rl}.


\clearpage
\section{Limitations}
The proposed system serves as a proof of concept demonstrating that \glspl{llm} can be trained through interaction with its environment. While the results indicate that this approach is promising, several limitations remain.
First, the current system does not scale well with the trend toward massive parallelization in robotic \gls{rl} training \cite{massive_parallel_rl}. In traditional \gls{rl} settings, for example, with lightweight locomotion \glspl{nn}, parallel robot simulation can be exploited, due to the assumption of \gls{nn} inference being significantly faster than simulation, allowing the \gls{rl} policy to learn from many concurrent robot observations. However, integrating \glspl{llm} into an \gls{rl} loop introduces substantial computational costs, making inference longer and, as such, simulation parallelization less effective.
Second, the simulation environment here is based on \gls{ros} and tightly coupled with an \gls{mpc} controller, which can become brittle during extended training runs. In particular, instability or crashes in the \gls{mpc} solver can halt training, making the training process challenging.
Finally, while the proposed framework is in principle applicable to other robotic systems where \glspl{llm} can interact with a high-level controller, the current implementation is tailored specifically to an autonomous car platform. Adapting it to other robotic domains would require integration work and system-specific adaptations.



\bibliography{references}  

\appendices
\section*{Appendix}
\section{Training Curves} 
\begin{figure*}[!htb]
    \centering
    \begin{subfigure}[b]{0.49\textwidth}
        \centering
        \includegraphics[height=5cm]{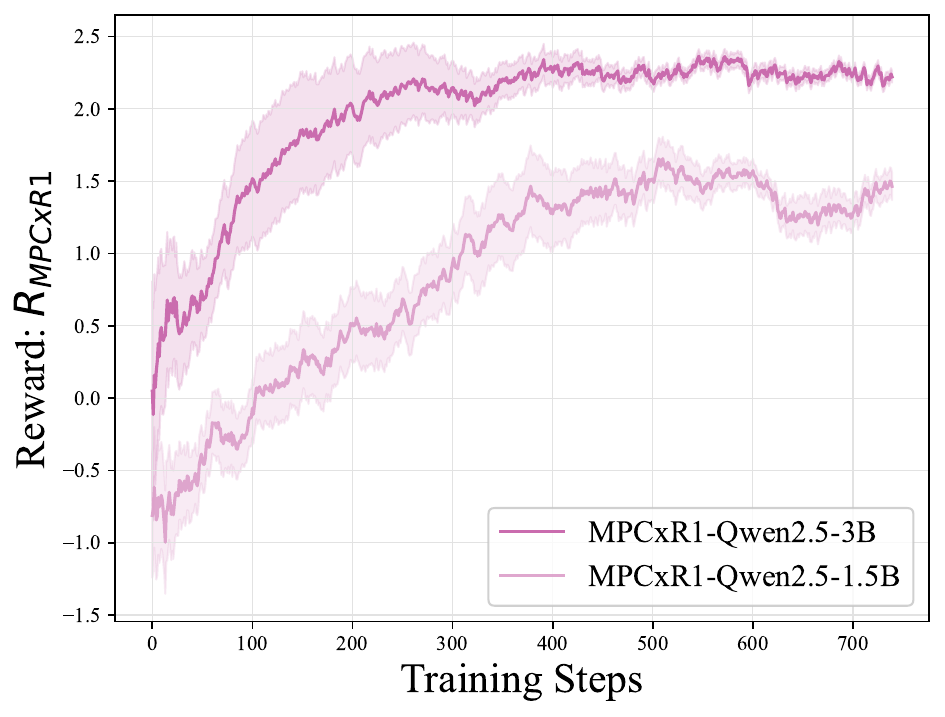}
        \label{fig:train_rew_mpc}
    \end{subfigure}
    \hfill
    \begin{subfigure}[b]{0.49\textwidth}
        \centering
        \includegraphics[height=5cm]{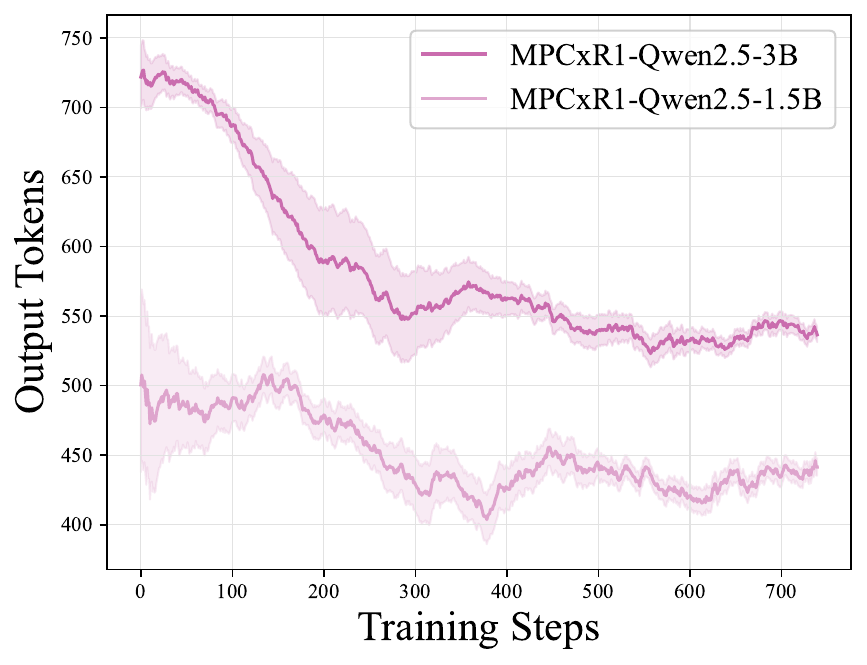}
        \label{fig:train_tks_mpc}
    \end{subfigure}
    \caption{Visualization of \textit{MPCxR1} \gls{rlvr}
    (\gls{grpo}) training with the standard deviation shaded. \textit{Qwen2.5} 3B and 1.5B \glspl{llm}, both pretrained via \gls{sft}, are used as base models. Left: The reward signal $R_{\text{MPCxR1}}$ that the \gls{llm} learns to maximize through interaction with the simulation environment. Right: The average output token length.}
    \label{fig:train_mpc_curves}
\end{figure*}

\begin{figure*}[!htb]
    \centering
    \begin{subfigure}[b]{0.49\textwidth}
        \centering
        \includegraphics[height=5cm]{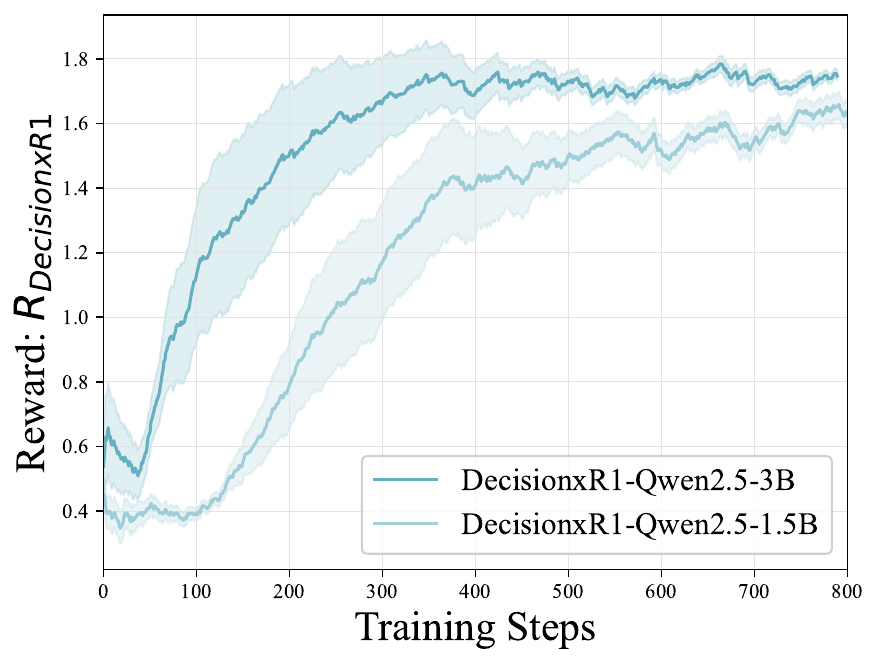}
        \label{fig:train_rew_decision}
    \end{subfigure}
    \hfill
    \begin{subfigure}[b]{0.49\textwidth}
        \centering
        \includegraphics[height=5cm]{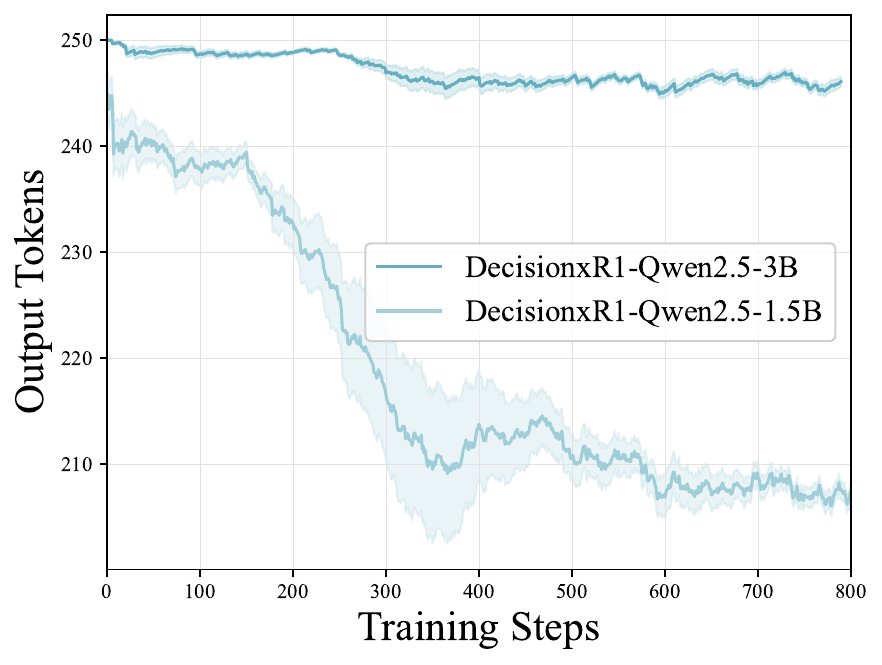}
        \label{fig:train_tks_decision}
    \end{subfigure}
        \caption{\textit{DecisionxR1} \gls{rlvr}
    (\gls{grpo}) training curves. Left: Reward $R_{\text{DecisionxR1}}$ learned from static binary adherence classification. Right: Output token length over the training steps.}
    \label{fig:train_decision_curves}
\end{figure*}

\label[appendix]{app:training_curves}
This section presents the \gls{rlvr} training curves for both the \textit{MPCxR1} and \textit{DecisionxR1} modules, shown in \Cref{fig:train_mpc_curves} and \Cref{fig:train_decision_curves}, respectively. The models used in both training setups are \textit{Qwen2.5-1.5B} and \textit{Qwen2.5-3B}, each pretrained via \gls{sft}. The training runs correspond to the final two rows in \Cref{tab:control_res} and \Cref{tab:decision_res}, where performance is evaluated after both \gls{sft} and \gls{rlvr} training.

The left plots illustrate the total reward ($R_{\text{MPCxR1}}$ and $R_{\text{DecisionxR1}}$) over training steps, showing consistent reward maximization in both cases. In \textit{MPCxR1}, the reward emerges from closed-loop interactions with the simulator, whereas in \textit{DecisionxR1}, it is learned from static dataset supervision.

The right plots show the average output token length over training. Interestingly, in both cases, the \gls{llm} learns to produce shorter, more concise outputs, despite the fact that brevity is not explicitly incentivized through the reward function. This contrasts with the behavior observed in \emph{DeepSeek R1} \cite{r1}, where longer \glspl{cot} were rewarded for multi-step mathematical reasoning. These results suggest that the reasoning tasks explored in this work differ from mathematical settings, and align with recent findings in spatial reasoning with \glspl{vlm} trained using \gls{rlvr} on static datasets \cite{embodiedR}.

\section{MPCxR1 Training and Evaluation Map}
\begin{figure}[htb]
    \centering
    \begin{adjustbox}{center}
        \includegraphics[width=0.8\textwidth]{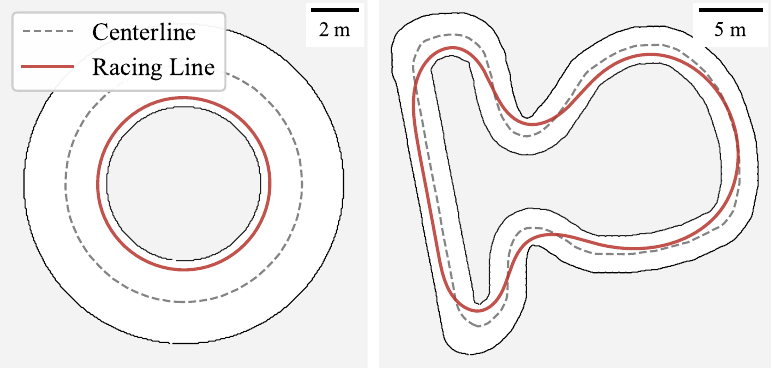}
    \end{adjustbox}
    \caption{Tracks used for training and evaluation. Left: \textbf{Circle} map used for \emph{MPCxR1} \gls{rlvr} training. Right: \textbf{The Grand Tour} map used for evaluation. Dashed grey and solid red lines denote the centerline and the minimum-curvature racing line (targeted by the default \gls{mpc}), respectively.}
    \label{fig:tracks}
\end{figure}

\label[appendix]{app:maps}
The maps used in the proposed \textit{MPCxR1} \gls{rlvr} training and evaluation procedure are shown in \Cref{fig:tracks}. The \textbf{Circle} map, used during training, provides a minimal and structured environment in which the \gls{llm} can efficiently learn to interact with the robot and environment. In contrast, \textbf{The Grand Tour} map, used for control adaptability evaluation as in \Cref{tab:control_res}, follows a conventional racing circuit design with greater complexity and variation.

This deliberate separation between training and evaluation environments is intended to assess the generalization capabilities of the learned driving behaviors. By training on a simple layout and testing on a more challenging and realistic track, the system's ability to transfer learned control adaptations across domains is effectively demonstrated.

\section{Behavioral Sets in Training and Evaluation}\label[appendix]{app:thebss}

To give more insight into the behavioral set used during \gls{rlvr} training of \textit{MPCxR1}, we provide the list of behaviors the \gls{llm} is trained to adapt toward. These behaviors form the training set $\mathcal{B}$ and are evaluated on the simple \textbf{Circle} track. Importantly, the reference \gls{rmse} values ($E^{\text{MPC}}_\mathcal{B}$) presented below correspond specifically to this training map. As such, they differ from the values reported in \Cref{tab:control_res}, where evaluation was conducted on \textbf{The Grand Tour} map, a more complex and realistic evaluation track.

\bm{$\mathcal{B}$} during training on the \textbf{Circle map} consists of:
\begin{enumerate}[I]
    \item \textbf{Centerline tracking:} Prompt: \emph{"Drive on the centerline"}. The \gls{rmse} evaluation $E_C$ computes the distance toward the centerline over a lap. The default \gls{mpc} tracks a minimum-curvature racing line, deviating from the centerline, yielding $E^{\text{MPC}}_C = 1.23$.
    
    \item \textbf{Velocity adherence:} Prompt: \emph{"Drive at 1.83 m/s as closely as possible"}. The \gls{rmse} evaluation $E_V$ computes the deviation from the reference velocity of \SI{1.83}{\metre\per\second}. As the default \gls{mpc} aims for time-optimal speed, this yields $E^{\text{MPC}}_V = 2.1$.
    
    \item \textbf{Reversing:} Prompt: \emph{"Reverse the car"}. The \gls{rmse} evaluation $E_R$ computes the deviation from \SI{-1}{\metre\per\second} (negative velocity indicating reverse motion), where $E^{\text{MPC}}_R = 4.93$.
    
    \item \textbf{Smooth driving:} Prompt: \emph{"Drive smoothly"}. The \gls{rmse} evaluation $E_S$ measures deviation from \SI{0}{\metre\per\second\squared} acceleration. Due to the uniform curvature of the circular track, the default \gls{mpc} exhibits low acceleration, resulting in $E^{\text{MPC}}_S = 0.38$.
\end{enumerate}

The randomized and perturbed behavioral set \bm{$\mathcal{B}'$} during evaluation, which are unseen during \gls{sft} and \gls{rlvr} training, on the \textbf{Grand Tour map} consists of. Here, the mean of each $E^{\text{MPC}}_{\mathcal{B}'}$ corresponds to the \gls{mpc} default parameters in \Cref{tab:control_res}: 
\begin{enumerate}[i]
    \item \textbf{Centerline tracking:} Prompts: [\emph{"Stay directly on the middle of the track"}, \emph{"Follow the track by staying aligned with the middle of the track"}, \emph{"Drive away as far as possible from the walls"}, \emph{"Ensure that the distance to the left and right wall remain the same"}, \emph{"Drive on the centerline"}]. The \gls{rmse} evaluation $E_C$ computes the distance toward the centerline over a lap. The default \gls{mpc} tracks a minimum-curvature racing line, deviating from the centerline, yielding $E^{\text{MPC}}_C = 0.679$, which is the same for all 5 prompts.
    
    \item \textbf{Velocity adherence:} Prompts: [\emph{"Set the driving speed to 3.5 m/s"}, \emph{"Target a driving speed of 2.2 meters per second"}, \emph{"Move at a constant speed of 1.25 m/s"}, \emph{"Travel at 2.9 meters per second"}, \emph{"Adjsut the speed to exactly 4.5 m/s"}]. The \gls{rmse} evaluation $E_V$ computes the deviation from the reference velocity of [3.5, 2.2, 2.25, 2.9, 4.5] \si{\metre\per\second} respectively. As the default \gls{mpc} always aims for the time-optimal speeds with respect to a range of reference velocities, this yields $E^{\text{MPC}}_V =$ [1.362, 2.417, 3.301, 1.808, 1.045] respectively.
    
    \item \textbf{Reversing:} Prompts: [\emph{"Slowly back the vehicle up"}, \emph{"Reverse the vehicle"}, \emph{"Switch to reverse and drive backwards"}, \emph{"Retreat by reversing the car"}, \emph{"Drive the car backwards"}]. The \gls{rmse} evaluation $E_R$ computes the deviation from \SI{-1}{\metre\per\second} (negative velocity indicating reverse motion), where $E^{\text{MPC}}_R = 5.439$, which is the same for all 5 prompts.
    
    \item \textbf{Smooth driving:} Prompts: [\emph{"Drive in a fluid and controlled manner"}, \emph{"Maintain low jerk and high smoothness in driving behavior"}, \emph{"Focus on smooth driving"}, \emph{"Drive smoothly"}, \emph{Reduce lateral acceleration"}]. The \gls{rmse} evaluation $E_S$ measures deviation from \SI{0}{\metre\per\second\squared} acceleration. The default \gls{mpc} exhibits rather high acceleration levels due to the non-uniform track layout, resulting in $E^{\text{MPC}}_S = 1.773$.
\end{enumerate}
    
\section{Qualitative LLM Outputs} \label[appendix]{app:car_bubbles}
This is the chat log of the experiment conducted on the physical robot in \Cref{fig:car_experiment}. Initially, the robot has been manually configured to oscillate (by lowering the heading trackign cost \texttt{qalpha} to 0, and by over inflating the \texttt{track\_safety\_margin} to \SI{1}{\metre} on a narrow and edgy track), then \textit{MPCxR1} is tasked to drive smoothly at \SI{2}{\metre\per\second}, followed by a command to reverse the car and then to drive nominally again. The prompt bubble's are omitted after the first one for visibility reasons.

Note that the computation times within this experiment might differ from the results obtained in \Cref{app:compute}, as the \gls{llm} inference was now performed with the entire autonomy stack running in parallel, which is computationally expensive. 

\begin{mpcbubble}{2a}{Instruction}
    \emph{"Drive smoothly at 2m/s"}
\end{mpcbubble}

\begin{mpcbubble}{2b}{Prompt}
Adapt the tuneable parameters of the MPC so that the car achieves the following: \textbf{\textit{"Drive smoothly at 2m/s"}}.\\
This is the MPC formulation: \texttt{MPC Formulation}.\\
Return format:
\vspace{-0.15cm}
\begin{verbatim}
new_mpc_params = {
    param1: new_value1,
    param2: new_value2,
    ...
}
\end{verbatim}
\end{mpcbubble}

\begin{mpcbubble}{2c}{Response}
\texttt{new\_mpc\_params} = \texttt{\{'qv': 10, 'qn': 20, \textbf{'qalpha': 7}, 'qac': 0.01, 'qddelta': 0.1, \textbf{'v\_max': 2.0}, 'v\_min': 1.0\}}\\
\end{mpcbubble}

\begin{narratebubble}
The robot was initially oscillating heavily, as in \Cref{fig:car_experiment}, due to the manually induced behavior of setting \texttt{qalpha} to zero and by inflating overly \texttt{track\_safety\_margin} to \SI{1}{\metre}. The \emph{MPCxR1} correctly limits the velocity to \SI{2}{\metre\per\second}, and adapts \texttt{qalpha} to 7. The robot is now driving smoothly at \SI{2}{\metre\per\second} on the racing line. \textbf{Computation time} on the Jetson Orin AGX: \textbf{\SI{9.12}{\second}}.
\end{narratebubble}

\begin{mpcbubble}{2d}{Instruction}
    \emph{"Reverse the car"}
\end{mpcbubble}


\begin{mpcbubble}{2e}{Response}
\texttt{new\_mpc\_params} = \texttt{\{\textbf{v\_max: -1, v\_min: -2}\}}\\
\end{mpcbubble}

\begin{narratebubble}
The robot was tracking the racing line, when prompted to reverse. The \textit{MPCxR1} module correctly identified the relevant parameters that enforce reversing behavior, by only allowing the \gls{mpc} to operate within negative velocity constraints. The robot is now reversing. \textbf{Computation time} on the Jetson Orin AGX: \textbf{\SI{5.84}{\second}}.
\end{narratebubble}

\begin{mpcbubble}{2f}{Instruction}
    \emph{"Drive normal again"}
\end{mpcbubble}


\begin{mpcbubble}{2g}{Response}
\texttt{new\_mpc\_params} = \texttt{\{'qv': 10, 'qn': 20, 'qalpha': 7, 'qac': 0.01, 'qddelta': 0.1, 'alat\_max': 10, 'a\_min': -5, 'a\_max': 5, 'v\_min': 1, 'v\_max': 5, 'track\_safety\_margin': 0.45\}.}\\
\end{mpcbubble}

\begin{narratebubble}
The robot is now driving normally again by using the default \gls{mpc} parameter values. \textbf{Computation time} on the Jetson Orin AGX: \SI{9.93}{\second}
\end{narratebubble}

\setcounter{bubblegroup}{2}

\section{Additional RAG Information}\label[appendix]{app:rag}
This section provides supplementary details on the \gls{rag} information used in both the \emph{DecisionxR1} and \emph{MPCxR1} modules. The design and usage closely follow the implementation presented in \cite{rsspaper}, Section III-B. This appendix is intended to offer additional insight into the structure and purpose of the retrieved context used to support reasoning and adaptation during training and inference.

The \glspl{rag} of \Cref{lst:mpc_rag} and \Cref{lst:decision_rag}, are meant to augment the prompts for their respective \textit{MPCxR1} and \textit{DecisionxR1} modules, respectively. This is a simple way in which a human operator can quickly integrate new system knowledge into the \gls{llm}. For example, if one were to deploy the proposed system on a full-scale car, changing the nominal speeds in \Cref{lst:decision_rag} higher than those observed on a 1:10 scaled racing car, would be as simple as editing \texttt{Hint 2} in \Cref{lst:decision_rag} through the \gls{rag} \texttt{.txt} file.

\subsection{RAG for Controller Adaptation}
\begin{lstlisting}[style=ragstyle, caption={RAG Memory for \textit{MPCxR1}. There are 11 memories in total.}, label={lst:mpc_rag}]
# Memory Entry 0:
Scenario:
To force going forwards v_min should be positive. If you want it to be able to reverse, then set v_min to negative.
MPC Action:
mpc_params = {
    'v_min': positive, if you want to go forwards, else negative to reverse
}

# Memory Entry 1:
Scenario:
Always have v_max be higher than v_min.
MPC Action:
mpc_params = {
    'v_max': higher than v_min
}

# Memory Entry 2:
Scenario:
To drive far away from the walls, the boundary inflation should be increased. That way it will drive closer to the centerline. This is the most important to stay away from the walls.
MPC Action:
mpc_params = {
    'track_safety_margin': high
}

# Memory Entry 3:
Scenario:
To reverse the car slowly on the racing line, the v_min must be negative!
MP Action:
mpc_params = {
    'v_min': negative,
}
...
\end{lstlisting}

\subsection{RAG for Decision-Making}

\begin{lstlisting}[style=ragstyle, caption={RAG Memory for \textit{DecisionxR1}. There are 11 hints in total.}, label={lst:decision_rag}]
# Hint 1:
If the d-speed is above than 0.5m/s is high.

# Hint 2:
Unless specified differently by the human, the car is usually driving at speeds between 5 and 7m/s.

# Hint 3:
If the distance to a wall is smaller than 0.4m, the car is close to that wall. Staying close to the wall means maintaining a consistent distance below 0.4m.

# Hint 4:
If the s-speed is close to 0m/s, then the car is stopped.

# Hint 5:
The car is reversing if the s-speed is negative or if the s-coordinate is reducing over multiple samples.
...
\end{lstlisting}

\section{Computation}\label[appendix]{app:compute}
This section analyzes the computational characteristics of the proposed training and deployment pipeline, focusing primarily on model quantification and its impact on both performance and deployability on embedded hardware. To assess the trade-offs between reasoning capability and computational efficiency in real-world robotic applications, we evaluated different model sizes, precision levels, and deployment environments. 

To reduce inference latency, we quantized the \gls{sft} and \gls{rlvr} trained \glspl{llm} by merging the \gls{lora} adapters into the base model and applying post-training quantization, which converts the models to a \texttt{Q5\_k\_m} quantized model, denoted as Q5, and exported in \texttt{GGUF} format and run using the \texttt{llama.cpp} inference engine \cite{llamacpp}. Full-precision models are here denoted as FP16, while quantized variants are referred to as Q5.

\begin{table}[htb] 
    \centering 
    \small	
    \setlength{\tabcolsep}{0.7 mm}
    \begin{tabular}{l|l|c|c|c|c|c|c}
    \toprule
    \textbf{HW} & \textbf{LLM} & \textbf{Quant} & \textbf{Param} & \textbf{Mem} & \textbf{Tokens} & \textbf{Tokens/s} & \textbf{Latency [s]} \\
    & & & \textbf{[\#B]} & \textbf{[GB]} & \textbf{[\#]} & \textbf{[s$^{-1}$]$\uparrow$} & $\mu_{t}$  \\
    \midrule
    &  & & 1.5 & 2.7 & 104 & \textbf{33.88} & \textbf{3.07}  \\
     & Qwen  & FP16 & 3 & 4.4 & 109 & 18.03 & 6.05 \\
    RTX &  &  & 7 & 8.4 & 108 & 8.7 & 12.41  \\
    \cmidrule{2-8}
    3090 &  &  & 1.5 & 1.7 & 102 & \textbf{203.89} & \textbf{0.50}  \\
    & Qwen & Q5 & 3 & 2.8 & 137 & 141.57 & 0.97 \\
    &  &  & 7 & 5.6 & 40 & 100.74 & 0.40 \\
    \midrule
    \multirow{6}{*}{\tabincell{l}{Jetson\\Orin\\AGX}} &  & & 1.5 & 2.2 & 104 & \textbf{6.99} & \textbf{14.88}  \\
    & Qwen  & FP16 & 3 & 3.8 & 109 & 3.55 & 30.73 \\
    &   & & 7 & 6.9 & 108 & 1.60 & 67.45 \\
    \cmidrule{2-8}
    & &  & 1.5& 1.6 & 102 & \textbf{56.14} & 1.82  \\
    & Qwen  & Q5 & 3 & 2.7 & 145 & 38.78 & 3.74 \\
    &  &  & 7 & 5.3 & 40 & 23.36 & \textbf{1.71}  \\
    \bottomrule
    \end{tabular}
    \vspace{4pt}
    \caption{Comparison of computational performance of quantized and full-precision models. The \glspl{llm} were deployed on both an RTX 3090 GPU and the \textit{Jetson Orin AGX} robotic \gls{obc}. The number of tokens denotes the output tokens generated for the given inference. The average inference latency is denoted with $\mu_{t}$.}
    \label{tab:compute}
\end{table}

To quantify the inference speedup gained by quantization and the optimized inference engine \cite{slm, llamacpp}, we measured token throughput, inference latency, and memory usage of each model using the same input prompt. Each measurement was performed 10 times sequentially and conducted on two hardware platforms: an RTX 3090 GPU, representing high-end consumer hardware, and a Jetson Orin AGX, targeting onboard deployment. 

The results are summarized in \Cref{tab:compute} and show that quantized models offer considerable gains in throughput, especially on embedded hardware. On the Jetson AGX, for instance, the Q5 1.5B model achieves a latency of 1.82 s and a throughput of 56.14 tokens/s, compared to the FP16 version's 14.88 s latency and 6.99 tokens/s. This corresponds to a 7.5x reduction in latency, making real-time inference feasible on embedded hardware. Similar trends are also observed in the other model sizes. Note that latency is determined through the token throughput and the number of output tokens produced, which is not the same for each model.

We further evaluated the impact of post-training quantization on the reasoning capabilities of our newly trained models, with a particular focus on decision-making and control adaptability. \Cref{tab:decision_res_app} displays how quantization affects decision accuracy across different model sizes. While full-precision models (FP16) consistently outperform their quantized counterparts, the accuracy drop due to Q5 quantization remains moderate. The largest decrease in accuracy is observed for the \textit{Qwen2.5-3B} model, where performance drops from 86.82\% to 79.88\%. Nevertheless, the quantized \gls{sft} \& \gls{rlvr}-trained model still outperforms the base model and other training configurations. This suggests that the reasoning capabilities learned through \gls{rlvr} remain largely intact in quantized models. As a result, the quantization of \gls{rlvr}-trained models leads to meaningful gains in speed and memory efficiency with only minor reductions in decision accuracy, making real-time onboard inference feasible. 

\begin{table}[htbp]
    \centering
    \small
    \setlength{\tabcolsep}{0.7 mm}
    \begin{tabular}{l|c|c|c|c}
    \toprule
    \textbf{LLM} & \textbf{Quant} & \textbf{SFT} & \textbf{GRPO} & \textbf{Accuracy [\%]}\bm{$\uparrow$}\\
    \midrule
    Qwen2.5-7B & FP16  & \cmark & \xmark & \textbf{87.32} \\
    \midrule
    Qwen2.5-1.5B & FP16 & \cmark & \cmark & 82.83\\
    Qwen2.5-3B & FP16 & \cmark & \cmark & \textbf{86.82}\\
    \midrule
    \midrule
    Qwen2.5-7B & Q5  & \cmark & \xmark & \textbf{86.95} \\
    \midrule
    Qwen2.5-1.5B & Q5 & \cmark & \cmark & 79.63\\
    Qwen2.5-3B & Q5 & \cmark & \cmark & \textbf{79.88}\\
    \bottomrule
    \end{tabular}
    \vspace{4pt}
    \caption{Impact of quantization on decision-making accuracy across multiple LLM sizes and training configurations.}
    \label{tab:decision_res_app}
\end{table}

\Cref{tab:quant_control_res_app} highlights that control adaptation, unlike decision making, places greater demands on the model, requiring it to generate valid, structured parameters that directly influence low-level control. This complexity makes the task more sensitive to reasoning stability, particularly under quantization.

The quantized 1.5B model, while achieving a solid improvement of 49.1\%, exhibits multiple extraction failures. These errors suggest that its reasoning capacity is insufficient to meet the stricter requirements of closed-loop control consistently. In contrast, the \textit{Qwen2.5-3B} Q5 model delivers more reliable performance, achieving 47.9\% improvement without any failures across all prompts.

This robustness under quantization makes the 3B model a better candidate for embedded deployment, as it strikes a balance between computational efficiency and control adaptability without sacrificing correctness.

\begin{table*}[htbp] 
    \centering 
    \begin{adjustbox}{center, max width=0.9\textwidth}
    \begin{tabular}{l|c|c|c|c|c|c|c||c|c}
    \toprule
    \textbf{LLM} & \textbf{Quant} & \textbf{SFT} & \textbf{GRPO} & \bm{$E_C [m]\downarrow$} & \bm{$E_V [ms^{-1}]\downarrow$} &  \bm{$E_R [ms^{-1}]\downarrow$} & \bm{$E_S [ms^{-2}]\downarrow$} & \textbf{Ext. Fail [\#]}\bm{$\downarrow$} & \textbf{Improve [\%]}\bm{$\uparrow$} \\
    \midrule
    MPC (default) & - & - & - & 0.68 & 1.98 & 5.44 & 1.77 & - & - \\
    \midrule
    Qwen2.5-7B & FP16 & \cmark & \xmark & 0.53 (21.5\%) & 0.24 (87.9\%) & \textbf{0.11 (97.9\%)} & 1.34 (24.5\%) & \textbf{0} & 58.0\% \\
    \midrule
    Qwen2.5-1.5B & FP16 & \cmark & \cmark & 0.62 (8.5\%) & 1.05 (47.0\%) & 0.72 (86.7\%) & 1.36 (23.6\%) & \textbf{0} & 41.5\% \\
    Qwen2.5-3B & FP16 & \cmark & \cmark & \textbf{0.41 (39.9\%)} & \textbf{0.19 (90.2\%)} & 0.48 (91.2\%) & \textbf{1.21 (31.8\%)} & \textbf{0} & \textbf{63.3\%} \\
    \midrule
    \midrule
    Qwen2.5-7B & Q5 & \cmark & \xmark & \textbf{0.57 (16.3\%)} & 0.90 (54.6\%) & \textbf{0.12 (97.8\%)} & 1.48 (16.6\%) & \textbf{0} & 46.4\%\\
    \midrule
    Qwen2.5-1.5B & Q5 & \cmark & \cmark & 0.59 (13.4\%)  & \textbf{0.65 (67.2\%)} & 1.18 (78.3\%)$\dagger$  & 1.10 (37.9\%)$\dagger$ & 3 & \textbf{49.1\%$\dagger$} \\
     Qwen2.5-3B & Q5 & \cmark & \cmark & 0.65 (3.7\%) & 0.93 (53.1\%)  & 0.32 (94.1\%) & \textbf{1.05 (40.8\%)}  & \textbf{0} & 47.9\%  \\
    \bottomrule
    \end{tabular}
    \end{adjustbox}
    \caption{Control adaptability evaluation of \texttt{Q5\_k\_m} quantized models (Q5) compared to their full-precision (FP16) counterparts.  Metrics include deviation from the centerline ($E_C$ [m]), reference velocity tracking error ($E_V$ [ms$^{-1}$]), reversing accuracy ($E_R$ [ms$^{-1}$]), and driving smoothness ($E_S$ [ms$^{-2}$]). \gls{mpc} parameter extraction failures (Ext. Fail [\#]), are where the \gls{llm} output could not be parsed due to invalid parameters. Percentage improvements are computed relative to the baseline \gls{mpc}, which tracks the racing line. The improvement column aggregates overall performance across all metrics. Each entry represents the mean over five independent runs using different prompts. $\dagger$ indicates averages computed excluding extraction failures.}
    \label{tab:quant_control_res_app}
\end{table*}

\end{document}